\documentclass[runningheads]{llncs}

\usepackage{eccv}

\usepackage{eccvabbrv}
\usepackage{graphicx}
\usepackage{booktabs}
\usepackage{wrapfig}
\usepackage{caption}
\usepackage[accsupp]{axessibility}  

\usepackage{hyperref}

\usepackage{orcidlink}

\usepackage{tabularx}
\usepackage{makecell}
\usepackage{tcolorbox}

\begin{document}
\newcommand{\coolname}{\textit{NSVS-TL}}

\title{Towards Neuro-Symbolic Video Understanding} 


\author{Minkyu Choi$^*$\inst{1,2}\orcidlink{0009-0007-6557-8865} \and
Harsh Goel$^\dagger$\inst{1,2}\orcidlink{0009-0006-9873-9584} \and
Mohammad Omama$^\dagger$\inst{1,2}\orcidlink{0009-0009-2402-4158} \and
\\
Yunhao Yang\inst{1}\orcidlink{0000-0002-7199-2508} \and
Sahil Shah\inst{1,2}\orcidlink{0009-0009-3640-2339} \and
Sandeep Chinchali\inst{1,2}\orcidlink{0000-0002-0601-3633}}
\authorrunning{M. Choi et al.}

\institute{The University of Texas at Austin TX 78712, USA \and
UT SWARM Lab, Department of Electrical and Computer Engineering\footnote{\href{https://utaustin-swarmlab.github.io}{https://utaustin-swarmlab.github.io}}
$^*$Corresponding author: {\tt \small minkyu.choi@utexas.edu}\\
$^\dagger$Contributed equally to this work.}

\maketitle
\begin{abstract} 
The unprecedented surge in video data production in recent years necessitates efficient tools to extract meaningful frames from videos for downstream tasks. Long-term temporal reasoning is a key desideratum for frame retrieval systems. While state-of-the-art foundation models, like VideoLLaMA and ViCLIP, are proficient in short-term semantic understanding, they surprisingly fail at long-term reasoning across frames. A key reason for this failure is that they intertwine per-frame perception and temporal reasoning into a single deep network. Hence, decoupling but co-designing the semantic understanding and temporal reasoning is essential for efficient scene identification. We propose a system that leverages vision-language models for semantic understanding of individual frames and effectively reasons about the long-term evolution of events using state machines and temporal logic (TL) formulae that inherently capture memory. Our TL-based reasoning improves the F1 score of complex event identification by $9-15$\%, compared to benchmarks that use GPT-4 for reasoning, on state-of-the-art self-driving datasets such as Waymo and NuScenes. The source code is available at GitHub\footnote{\href{https://github.com/UTAustin-SwarmLab/Neuro-Symbolic-Video-Search-Temporal-Logic}{https://github.com/UTAustin-SwarmLab/Neuro-Symbolic-Video-Search-Temporal-Logic}}.
\keywords{ Video Understanding  \and  Video Reasoning \and Neuro Symbolic AI \and Temporal Logic \and Formal Methods }
\end{abstract}

\section{Introduction}
\label{sec:intro}

There is a significant increase in video data production, with platforms such as YouTube receiving $500$ hours of uploads every minute. Additionally, autonomous vehicle companies such as Waymo generate $10$-$100$ TB \cite{waymo} of data, and worldwide security cameras record around $500$ PB \cite{securitycams} of data daily. Consequently, we require tools with sophisticated query capabilities to navigate this immense volume of video content. For instance, a query such as ``Find me all scenes where event A happened, event B did not occur, and event C occurs hours later'' requires advanced methods capable of long-term temporal reasoning. Surprisingly, we find that today’s state-of-the-art (SOTA) video and language foundation models fail to identify complex events, especially when key-frames are rare in a video or when the input query is complex (see \cref{fig:fig2_nsvs_vs_benchmark_videomodels}).

Our key insight is that these foundation models aggregate per-frame semantics into a latent vector from which precise scene identification is difficult, especially over long videos. Instead, this paper innovates a novel neuro-symbolic approach that composes neural perception with temporal logic reasoning. We use vision-language models for per-frame semantic understanding but employ state machines and temporal logic to reason about the long-term evolution of events since they inherently capture memory. 

\begin{figure}[tb]
  \centering
  \captionsetup{belowskip=-10pt}
  \includegraphics[height=6.5cm]{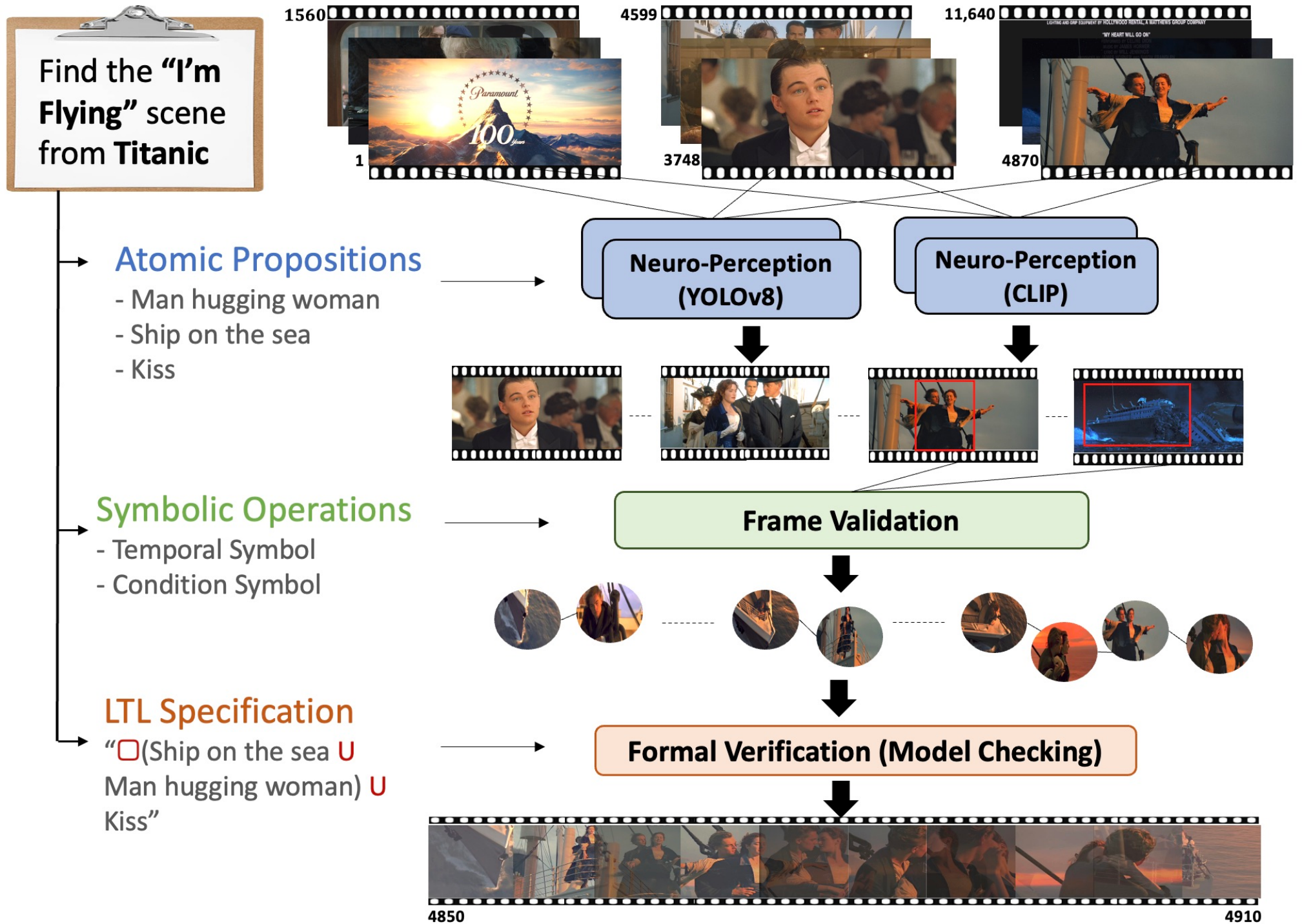}
  \caption{\textbf{\coolname \ Pipeline.} The input query --- ``Find the `I'm Flying scene from Titanic'' --- is first decomposed into semantically meaningful atomic propositions such as ``man hugging woman'', ``ship on the sea'', and ``kiss'' from a high-level user query. SOTA vision and vision-language models are then employed to annotate the existence of these atomic propositions in each video frame. Subsequently, we construct a probabilistic automaton that models the video's temporal evolution based on the list of per-frame atomic propositions detected in the video. Finally, we evaluate when and where this automaton satisfies the user's query. We do this by expressing it in a formal specification language that incorporates temporal logic. The TL equivalent of the above query is ALWAYS ($\Box$) ``man hugging woman'' UNTIL ($\mathsf{U}$) ``ship on the sea'' UNTIL ($\mathsf{U}$) ``kiss''. Formal verification techniques are utilized on the automaton to retrieve scenes that satisfy the TL specification.}
  \label{fig:teaser}
\end{figure}

Imagine we aim to pinpoint the iconic ``I’m flying'' scene in the movie Titanic. How do we identify a 5-minute-long scene precisely within the 3-hour and 14-minute movie? Our proposed method, Neuro-Symbolic Video Search with Temporal Logic (\coolname ), presents a solution using a neuro-symbolic approach for video understanding (see \cref{fig:teaser}). Our key contributions can be summarized as: 
\begin{enumerate} 
    \item \textbf{Real-time Neuro-Symbolic Scene Identification}: We propose a framework that segregates temporal reasoning from perception, overcoming the limitations of video-language models in localizing video segments in long videos. We employ perception models to identify frames relevant to the temporal logic query. Subsequently, we construct an automaton to verify the satisfiability of the temporal logic query. This segregation enables us to search for temporally extended scenes over long videos.
    \item \textbf{Formal Verification of  Video Frame Automaton against a TL Specification}: Our algorithm seamlessly provides a confidence measure for predictions on localized scenes by leveraging formal verification on a TL specification. We integrate the probabilistic model checker STORM \cite{junges2021stormpy} to formally verify the automaton constructed during scene identification and to measure the probability of satisfying the TL specification.
    \item \textbf{Datasets}: We propose Temporal Logic Video (TLV) datasets. These include a synthetic annotated video dataset created by stitching together images of objects from the ImageNet \cite{imagenet} and COCO \cite{coco} datasets to verify frame search algorithms. Additionally, we annotate autonomous vehicle datasets -- Waymo \cite{waymo} and NuScenes \cite{nuscenes} -- with TL specifications derived from ground truth object annotations such as ``bus'', ``pedestrian'', etc., in the scenes.
\end{enumerate}
\section{Related Work}
\label{sec:related_work}
\subsection{Video Search and Event Detection}
\begin{wrapfigure}{r}{0.5\textwidth}
    \centering
    \vspace{-30pt}
    \includegraphics[width=0.48\textwidth]{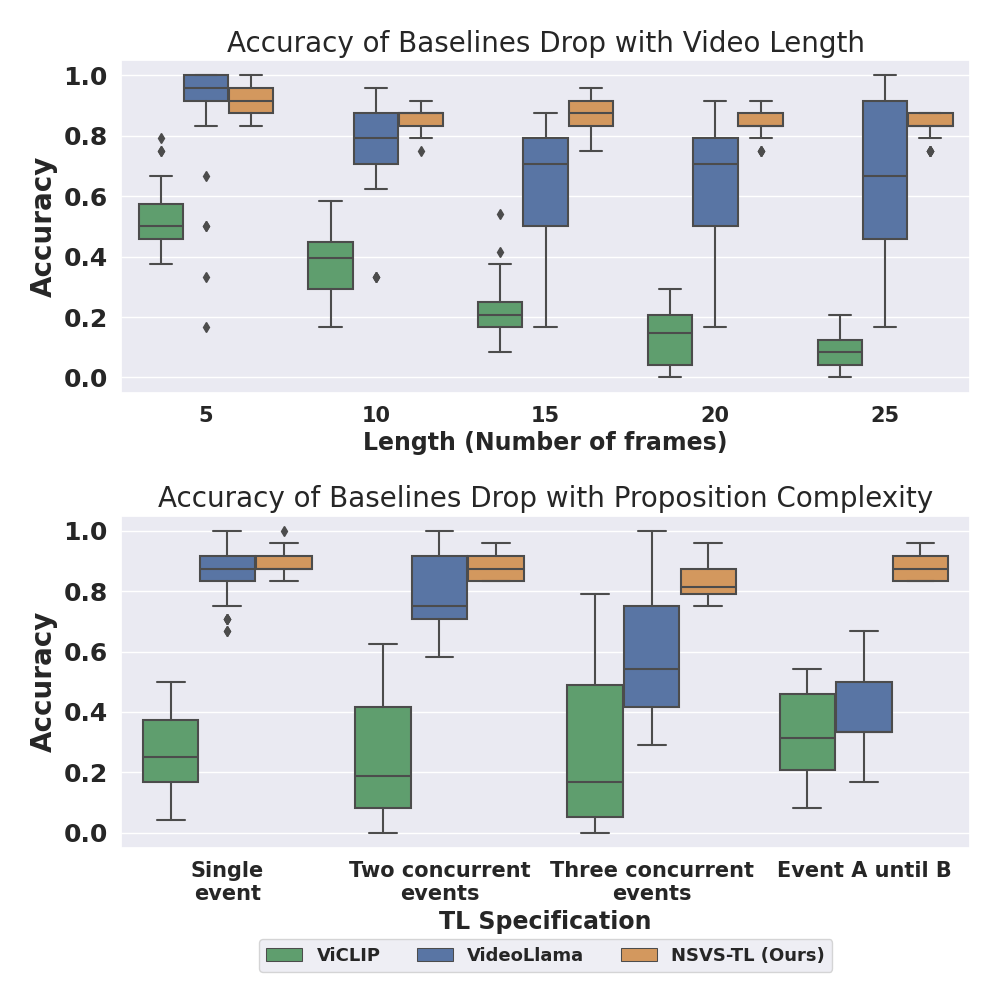} 
    \caption{\textbf{Comparative Performance on the Event Identification Task: Video Language Models versus \coolname.} The accuracy of event identification with Video Language Models (Blue/Green) drops as video length or query complexity increases. In contrast, \coolname \ (Orange) shows consistent performance irrespective of video length or query complexity.}
    \vspace{-25pt}
\label{fig:fig2_nsvs_vs_benchmark_videomodels}
\end{wrapfigure}
Existing research in video event detection predominantly focuses on tracking the spatio-temporal information of objects to identify events by employing deep neural networks for learning latent representations like objects' motion and position \cite{event-detect-spatiotemporal, event-detection-video-stream, lstm-event-detect, cnn-event-detect, zheng2022abnormal, interpretable-anomaly-detect, action-detect, action-detect-2}. These methods learn latent representations of videos for downstream event detection or classification, which require substantial computational resources for training neural networks \cite{lstm-event-detect, action-detect, action-detect-2}. In contrast, our approach leverages SOTA vision or vision-language models for per-frame semantic recognition, reducing the need for such infrastructure.

Additionally, some studies focus on searching and detecting events described in natural language \cite{interpretable-anomaly-detect, event-extract}. Video language models --- Video-Llama \cite{damonlpsg2023videollama} and Video-ChatGPT \cite{maaz2023video} --- integrate language foundation models like GPT-4 \cite{openai2023gpt4} and Llama \cite{touvron2023llama} for video question answering and zero-shot event recognition. However, their aggregation of temporal video information hinders accurate frame identification in long videos. To address this, some methods have introduced temporal logic for event detection \cite{temporal-event-detect, yang2023specification}. In contrast, our proposed method focuses on scene identification of any temporal length.

\subsection{Symbolic Video Understanding}
Many works explore approaches to building symbolic representations for video understanding. Such symbolic representations of videos are useful for video classification \cite{short1, short2}, event detection \cite{event-detection-video-stream, cnn-event-detect, lstm-event-detect}, video question-answering \cite{long1, neural-symbolic}, etc. Existing methods either construct graph structures \cite{graphical-model-relationship-detection, visual-symbolic, long3} or use latent-space representations as symbolic representations of videos \cite{symbolic-high-speed-video, BertasiusWT21, neural-symbolic-cv}.
Among the existing works, several of them propose methods of learning symbolic representation to understand long videos \cite{long1, long-form-video, long2, long3}. However, these approaches also rely on latent-space representations, which lack interpretability.  In contrast, we use a fully interpretable automaton-based representation that provides formal guarantees to downstream tasks like video search.

\section{Preliminaries}
\label{sec:preliminary}
We introduce a practical example to assist the reader in understanding our proposed methodology. Let's consider a scenario where our objective is to ensure that an autonomous vehicle follows a safety protocol when children are present in school zones. The key query for this test is: ``Find all scenes where the school zone sign is visible and children are present, continuing until the vehicle exits the school zone''.

\subsection{Neural Perception Model} \textit{Understanding the semantics of a frame}: 
The initial step involves identifying frames corresponding to propositions in the query --- such as ``school zone sign'' and ``children''. Therefore, we require the use of vision or vision-language neural perception models. Our method incorporates neural perception models to extract information from video frames, focusing on object detection capabilities. The confidence scores produced by these models are crucial as they indicate the level of certainty in detection results, which significantly impacts the accuracy of the subsequent temporal logic analysis and scene identification. 

\subsection{Temporal Logic}
\textit{Reasoning about events across time}: Temporal logic (TL) provides a structured framework for describing and reasoning about the temporal properties of sequences or processes \cite{Temporal-and-Modal-Logic, Manna}. It extends classical logic with temporal operators to express propositions about the flow of time. In video analysis, temporal logic can be used to define complex conditions or sequences of events within a video stream. The syntax of temporal logic is composed of a set of first-order logic operators, a set of temporal operators, and a set of propositions. The set of first-order logic operators includes AND ($\wedge$), OR ($\vee$), NOT ($\neg$), Implies ($\implies$), etc. The set of temporal operators includes Always ($\Box$), Eventually/Exist ($\diamondsuit$), Next ($\mathsf{X}$), Until ($\mathsf{U}$), etc. Intuitively, for events $a$ and $b$, $\Box a$ means $a = True$ for every step, $\diamondsuit b$ means there exists one step where $b=True$, and $a \mathsf{U} b$ means $a=True$ for all the steps before $b = True$, then $b=True$ for all the subsequent steps. A minimal syntax for temporal logic is specified as follows:
\begin{equation}
    \label{eq:(1)}
    \Phi ::= p_k \;|\; \neg \Phi \;| \; \Phi \;\vee \Phi\;| \; \Phi \;\wedge \Phi\; | \;\mathsf{X} \Phi\; | \;\Phi \; \mathsf{U}\; \Phi,
\end{equation}
where $\Phi$ is a temporal logic specification, $p_k \in P$ is an atomic proposition --- a statement or assertion that must be true or false. We denote $P$ as the set of all atomic propositions. Temporal logic includes linear temporal logic (LTL), computation tree logic (CTL), probabilistic computation tree logic (PCTL), etc. We present some of these terminologies in the Appendix. 

For instance, in our running example, the set of propositions $P$ includes atomic propositions $p_k$ such as ``SchoolZoneSign'' and ``Children'', and  the temporal logic specification for the query is formulated as:
\begin{equation}
    \label{eq:running_example_spec}
    \Phi = (\text{SchoolZoneSign} \; \wedge \;\text{children}) \; \mathsf{U}\; \neg \text{SchoolZoneSign}.
\end{equation}

\subsection{Probabilistic Automaton}
\textit{Constructing a model of video}: A \textit{Probabilistic Automaton} (PA) is a mathematical model used to represent dynamic systems transitioning across various states with associated probabilities in response to certain inputs. The PA is defined as a 5-tuple $\mathcal{A}=(Q, \Omega, \delta, Q_0, F)$, where:

\begin{itemize}
\item $Q$ is a finite set of states in which the automaton can reside.
\item $\Omega$ is a finite set of symbols, or inputs, that trigger transitions between states.
\item $\delta: Q \times \Omega \times Q \to [0,1]$ is the transition function that maps a transition from one state to another given an input symbol with the probability of that transition. The sum of the probabilities for all outgoing transitions from a state $q$ equals 1, ensuring a complete probability distribution.

\item $Q_0 \subset Q \land Q_0 \neq \emptyset$ is the set of initial states from which the automaton begins its operation.
\item $F \subseteq Q$ is the set of acceptance or terminal states representing successful terminations.
\end{itemize}

In the running example, $Q_0$ is defined by the initial detection of ``SchoolZoneSign'' and ``Children''. The terminal states $F$ might contain conditions where the ``SchoolZoneSign'' is no longer detected, indicating the vehicle's exit from the school zone. The PA is a sophisticated model that transitions through states based on the temporal logic specification $\Phi$ (see \cref{eq:running_example_spec}). The transition probabilities between these states are quantified by $\delta$, with $Q_0$ and $F$ denoting the start and end of the scenes of interest. $Q$ and \(\delta\) are illustrated in \cref{fig:running_example_automaton} in details. With this constructed PA, we now conduct formal verification to ascertain whether the PA conforms to our predefined specification $\Phi$.

\subsection{Formal Verification}
\textit{Verifying frames of interest against pre-defined specifications}: Formal verification involves mathematically proving that a system adheres to specified properties. One such technique under formal verification is \textit{Model Checking}, a formal verification technique leveraged to check if a model of a system adheres to a set of specified properties. Given a model $\mathcal{M}$ and a specification $\Phi$ expressed in a suitable logic, model checking systematically explores the state space of $\mathcal{M}$ to verify if $\Phi$ holds. If $\mathcal{M}$ satisfies $\Phi$, denoted as $\mathcal{M} \models \Phi$, the model checking algorithm returns true; otherwise, it returns false. In the context of our running example, this step ensures that the PA precisely adheres to the vehicle's compliance with safety protocols while driving in the school zone.
\section{Methodology}
\label{sec:methodology}
We introduce a novel way to identify scenes of interest using a neuro-symbolic approach. Given video streams or clips and a TL specification $\Phi$, Neuro-Symbolic Visual Search with Temporal Logic (\coolname)  identifies scenes of interest through four steps: 
\begin{itemize}
\item \textbf{Step 1}: We calibrate the confidence of neural perception models to ensure precise object detection. This calibration enables the detection of relevant propositions in a given frame to construct a PA.
\item \textbf{Step 2}: Subsequently, each frame undergoes a validation process with two distinct validation functions. This step ensures that only frames containing relevant visual information proceed to the next phase of the method.
\item \textbf{Step 3}: Upon validation, we construct a probabilistic automaton to encapsulate the temporal and logical relations between successive frames.
\item \textbf{Step 4}: Finally, we utilize model checking to determine whether a constructed automaton satisfies the TL specification. If an automaton passes this check, then a sequence of frames within the automaton is identified as a scene of interest by the given TL specification.
\end{itemize}

\paragraph{Calibrating Neural Perception Model.}
We first calibrate neural perception models, $f_{v}$, which detect atomic propositions $p_k$ in a frame $\mathcal{F}_t$, where $t$ represents the frame's time index with a confidence $\hat{y}$ for probabilistic verification. We use a generalized logistic function as a mapping estimation function $z(\hat{y})$, which calibrates the confidence of models into an accuracy metric, as defined in \cref{eq:mapping-est}:
\begin{equation} 
    \label{eq:mapping-est}
    z(\hat{y}) = \frac{1}{1 + e^{-k(\hat{y} - \hat{y}_0)}},
\end{equation}
where $k$ is a scaling factor that adjusts the sensitivity and \(\hat{y}_0\) is an inflection point. We optimize $k$ and \(\hat{y}_0\) to calibrate the neural perception models as presented in \cref{fig:fig4_calibration_plot}. We also empirically select optimal thresholds for true positives ($\gamma_{tp}$) and false positives ($\gamma_{fp}$). The calibration function, which incorporates these thresholds along with a mapping estimation function, is presented in \cref{eq:cal-eq}:
\begin{equation} 
    \label{eq:cal-eq}
    g(\hat{y};\gamma_{fp}, \gamma_{tp}) =
    \begin{cases}
    0 & \text{if } \hat{y} < \gamma_{fp}, \cr
    1 & \text{if } \hat{y} > \gamma_{tp}, \cr
    z(\hat{y}) & \text{otherwise}.
    \end{cases}
\end{equation}
For confidence below \(\gamma_{fp}\), it outputs $0$, indicating low reliability. Above \(\gamma_{tp}\), it outputs $1$, reflecting high accuracy. For intermediate confidence levels, it utilizes a mapping estimation $z(\hat{y})$ as illustrated in \cref{fig:fig4_calibration_plot}.  

\begin{figure}[tb]
  \begin{minipage}[t]{0.5\textwidth}
    \vspace{5pt} 
    \includegraphics[width=\textwidth]{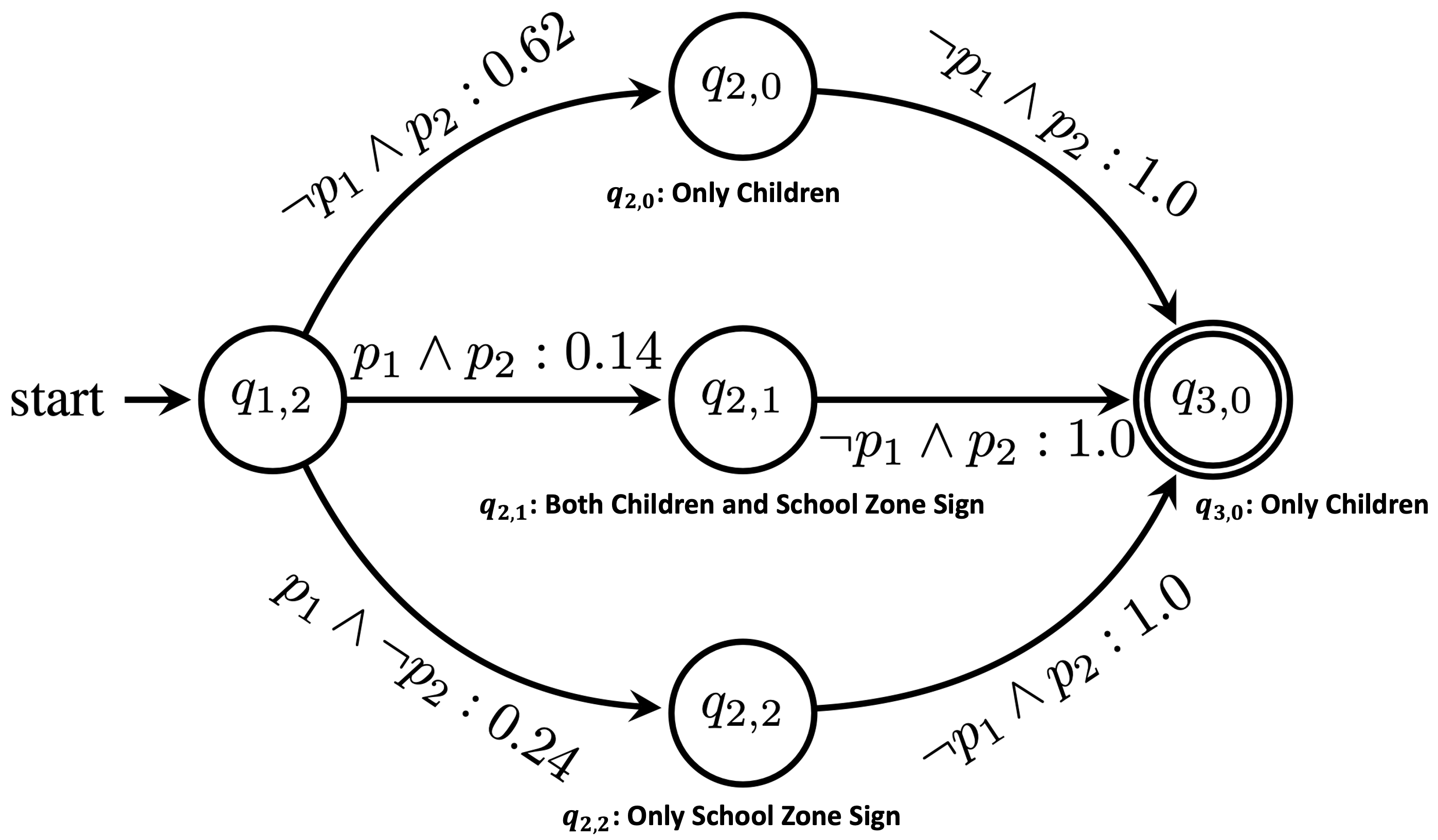}
  \end{minipage}\hfill
  \begin{minipage}[t]{0.45\textwidth}
    \caption{\textbf{Sample Automaton of the Running Example}. Illustrates transitions from \(\mathcal{F}_1\) to \(\mathcal{F}_3\). Key transitions include: $q_{1,2} \to q_{2,0}$ with a 0.62 probability for only children, $q_{1,2} \to q_{2,1}$ with a 0.14 probability for both sign and children, and $q_{1,2} \to q_{2,2}$ with a 0.24 probability for only the sign. There are only children in \(\mathcal{F}_3\). Therefore, all states in \(\mathcal{F}_2\) are connected to $q_3$ with 1.0 probability.}
    \label{fig:running_example_automaton}
  \end{minipage}
  \vspace{-2.0em} 
\end{figure}

\paragraph{Frame Validation.} 
Following the calibration step, the process transitions to frame validation. To ensure the presence of atomic propositions, we first validate a frame $\mathcal{F}_t$ defined by a detection verification function:
\begin{equation} 
    \label{eq:frame_val_cal}
    V_{c}(\mathcal{F}_t, P) = \begin{cases}
    1 & \text{if} \hspace{1mm} g(f_{v}(\mathcal{F}_t,p_k);\gamma_{fp},\gamma_{tp}) > 0, \forall p_k \in P\\
    0 & \text{otherwise}.
    \end{cases}
\end{equation}
A frame $\mathcal{F}_t$ is considered valid for the automaton if $V_c(\mathcal{F}_t,P) = 1$. However, this is not enough to consider the inclusion of \(\mathcal{F}_t\) in the automaton. For instance, consider a slight modification to our running example:
\begin{equation} 
    \label{eq:man-woman-animal}
    \Phi = (\text{SchoolZoneSign}  \wedge \text{children})  \mathsf{U} \neg \text{adults}.
\end{equation}  
In this case, we are interested in children walking around the school zone unaccompanied by adults. If we solely rely on the validation function from \cref{eq:frame_val_cal}, frames that include a school sign, children, and adults are added to the automaton. However, including frames with adults contradicts our requirement to exclude them. To address this, we implement \textit{symbolic verification} to ensure a frame \( \mathcal{F}_t \) aligns with our TL specification \( \Phi \). This process involves evaluating frames against both first-logic operators (denoted as \(\Psi\)) and temporal operators (denoted as \(\Theta\)) defined in \(\Phi\) as follows:
\begin{subequations}
  \begin{align}
    \Psi &= \{\psi \mid \psi \in \{\wedge, \vee, \neg, \ldots\} \land \psi \text{ appears in } \Phi\}, \\
    \Theta &= \{\theta \mid \theta \in \{\Box, \diamondsuit, \mathsf{U}, \ldots\} \land \theta \text{ appears in } \Phi\}.
    \label{eq:temporal_set}
  \end{align}
\end{subequations}
During a symbolic verification process, each frame initially evaluates the set of first-order logic operators \(\Psi\). Formally, \(\mathcal{F}_t\) is satisfied on \(\Psi\), denoted as $\mathcal{F}_t \models \Psi$, if and only if the following conditions hold:


\begin{itemize}
  \item \textbf{Conjunction} ($\wedge$): A frame $\mathcal{F}_t$ is satisfied on \(\Psi\) for the conjunction only if both propositions $p_1$ and $p_2$ associated by $\wedge$ are present in \(\mathcal{F}_t\).
  \item \textbf{Disjunction} ($\vee$): A frame $\mathcal{F}_t$ is satisfied on \(\Psi\) for the disjunction if at least one of the propositions $p_1$ and $p_2$ associated by $\vee$ is present in \(\mathcal{F}_t\). 
  \item \textbf{Negation} (\(\neg\)): A frame \(\mathcal{F}_t\) is satisfied on \(\Psi\) for the negation if the proposition \(p_k\) is not present in \(\mathcal{F}_t\). 
\end{itemize}

In our running example (see \cref{eq:man-woman-animal}), a frame \(\mathcal{F}_t\) must have a school zone and children, but not adults. This is represented by a combination of connection (\(\wedge\)) and negation (\(\neg\)) operators.  A frame that includes a school sign and children satisfies the conjunction requirement (\(\wedge\)). However, if adults are also present in the same frame, it violates the negation criterion (\(\neg\)), as our query explicitly requires the exclusion of adults.
\begin{figure}[tb]
  \begin{minipage}[t]{0.60\textwidth}
    \vspace{-5pt} 
    \includegraphics[width=\textwidth]{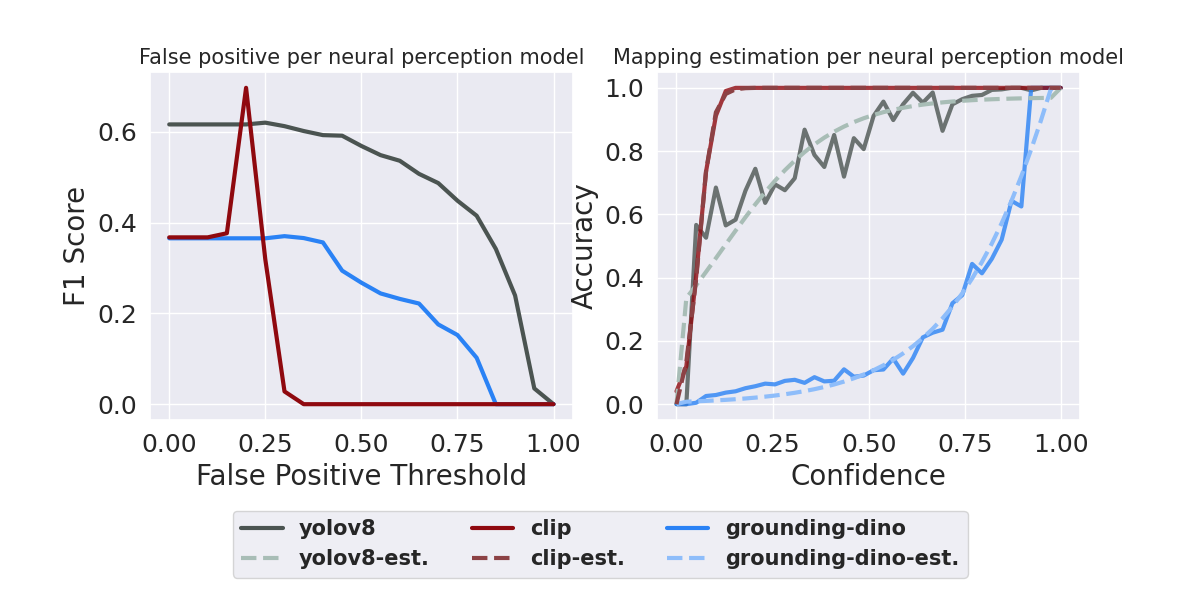}
  \end{minipage}\hfill
  \begin{minipage}[t]{0.45\textwidth}
    \vspace{0pt} 
    \caption{\textbf{Calibrating Neural Perception Models.} We empirically select the optimal false positive threshold as shown in the left figure, while the right figure illustrates mapping estimation functions optimized for each neural perception model.}
    \label{fig:fig4_calibration_plot}
  \end{minipage}
  \vspace{-1.5em}
\end{figure}

While the first part of symbolic verification, using logical operators (\(\Psi\)) helps identify frames where children are present without adults ($\neg \mathrm{adult}$), it can ignore the important temporal aspects of our query. Consider a scenario in which our objective is to observe children without adults in the school zone during the period before any adult arrives. If we only validate frames on \(\Psi\), a frame meets the immediate logical requirements (children without adults) yet fails to capture the essential temporal aspects --- the arrival of adults. This deficiency highlights the importance of integrating the temporal operators (\(\Theta\)) in our evaluation. 
Here, we examine the temporal relevance of each proposition in \(\Phi\) as follows:
\begin{equation}
    \label{eq:(10)}
    \tau(p_k) = 
    \begin{cases} 
    1 & \text{if } \exists \hspace{2mm} p_k \hspace{2mm} \text{such that} \hspace{2mm} \theta p_k,\\
    1 & \text{if } \exists \hspace{2mm} p_k, p_{k+1} \hspace{2mm} \text{such that} \hspace{2mm} p_k \hspace{0.4mm} \theta \hspace{0.7mm} p_{k+1}, \\
    0 & \text{otherwise},
    \end{cases}
\end{equation}
where \(\theta \in \Theta\) (see \cref{eq:temporal_set}). In conclusion, by the symbolic verification rules defined above, a frame $\mathcal{F}_t$ is validated by the symbolic verification function expressed as:
\begin{equation} 
    \label{eq:(11)}
    V_{sv}(\mathcal{F}_t, P) = \begin{cases}
    1 & \text{if} \hspace{1mm} (\exists \hspace{2mm} p_k \in P : \tau(p_k)=1) \vee (\mathcal{F}_t \models \Psi), \\
    0 & \text{otherwise}.
    \end{cases}
\end{equation}
Therefore, combining the symbolic verification function with the detection verification function (see \cref{eq:frame_val_cal}), the frame validation function is defined as follows:

\begin{equation} 
    \label{eq:final-validation-func}
    V(\mathcal{F}_t, P) = \begin{cases}
    1 & \text{if} \hspace{1mm} (V_{c}(\mathcal{F}_t,P) = 1) \wedge (V_{sv}(\mathcal{F}_t,P) = 1), \\
    0 & \text{otherwise}.
    \end{cases}
\end{equation}
A frame \(\mathcal{F}_t\) is valid for the automaton if $V(\mathcal{F}_t, P)$ = 1.

\paragraph{Dynamic Automaton Construction.} \label{par:pa}
Once a frame \(\mathcal{F}_t\) is validated by \cref{eq:final-validation-func}, we construct a PA as a model of the video with validated frames. We call this process \textit{Dynamic Automaton Construction} because we dynamically construct the automaton based on frame validation instead of building it for an entire video. We first generate $2^{|P|}$ new automaton states, where $|P|$ is the number of atomic propositions. Each of these states corresponds to a specific combination of propositions from the set $P$, thereby allowing a comprehensive representation of all possible combinations of propositions within the automaton. In our running example (see \cref{eq:running_example_spec}), we construct four states representing all possible combinations of atomic propositions or their negation with the proposition set $P =\{p_1,p_2\}$, where $p_1=\text{SchoolZoneSign}$ and $p_2=\text{Children}$. The labels for these states would be $p_1 \wedge p_2, \neg p_1 \wedge p_2, p_1 \wedge \neg p_2$, and $\neg p_1 \wedge \neg p_2$, respectively. 

After creating new states, we compute the transition probabilities from previous states to new states $q_{t,\omega}$ where $\omega \in 2^{|P|}$ as follows:

\begin{equation} 
    \label{eq:(13)}
    \mathbb{P}_{t,\omega} = \underset{p_k}{\mathrm{\prod}} g(f_{v}(\mathcal{F}_t,p_k);\gamma_{fp}, \gamma_{tp}) \hspace{2mm} \forall p_k \in P.
\end{equation}


The probability of the negation of a proposition ($\neg p_k$) is computed as $1-g(f_{v}(\mathcal{F}_t,p_k);\gamma_{fp}, \gamma_{tp})$. For every new state $q_{t,\omega}$ with a probability $\mathbb{P}_{t,\omega} > 0$, we construct transitions from the previous states $q_{{t-1},\omega}$ to $q_{t,\omega}$.
In this case, $\mathbb{P}_{t,\omega}$ represents the probability of the transition from any state in frame $\mathcal{F}_{t-1}$ to the new state $q_{t,\omega}$ of $\mathcal{F}_{t}$. We illustrate how the automaton is created for our running example in \cref{fig:running_example_automaton}.


\paragraph{Model Checking.} \label{para}
Once we have constructed the automaton (model of a video), we apply model checking using STORM \cite{stormpy}, a probabilistic model checking method, to determine whether it satisfies our specification formulated by probabilistic computation tree logic (PCTL). Our method iteratively processes frames and dynamically builds the automaton. This ensures that its evolving structure aligns with the desired TL specifications. This verification step checks the automaton's transitions and states to ensure they comply with the probabilistic temporal properties defined in the TL specification.

When the automaton \(\mathcal{A}\) (see \cref{sec:preliminary}) satisfies \(\Phi\),  it indicates that all frames in \(\mathcal{A}\) correspond to our defined scenes of interest. Subsequently, all frames representing the automaton's states are added to the scenes of interest set \(\mathcal{S}\). After this addition, the automaton \(\mathcal{A}\) is reset ($\mathcal{A} = \emptyset$), preparing the system for processing subsequent frames. The process then proceeds with frame \(\mathcal{F}_{t+1}\), repeating the previous steps and including frames that satisfy \(\Phi\) into \(\mathcal{S}\). In the context of our running example, all frames are added to \(\mathcal{S}\) when an autonomous vehicle exits a school zone where children are present. 
\section{TLV: Temporal Logic Video Datasets}
\label{sec:tlv_dataset}
Given the lack of SOTA video datasets for long-horizon, temporally extended activity and object detection, we introduce the Temporal Logic Video (TLV) datasets. The synthetic TLV datasets are compiled by stitching together static images from image datasets like COCO \cite{coco} and ImageNet \cite{imagenet}. This enables the artificial introduction of a wide range of TL specifications. Additionally, we create two video datasets annotated with TL specifications based on the open-source autonomous vehicle (AV) driving datasets NuScenes \cite{nuscenes} and Waymo \cite{waymo}. The dataset is available on GitHub\footnote{\href{https://github.com/UTAustin-SwarmLab/Temporal-Logic-Video-Dataset}{https://github.com/UTAustin-SwarmLab/Temporal-Logic-Video-Dataset}}.

\subsection{Dataset Compilation}
The synthetic TLV datasets are compiled by combining static images randomly selected from the COCO and ImageNet datasets. The propositions selected from the ground truth labels of the original datasets are used to generate TL specifications. For example, if `person' and `car' are selected from the original COCO dataset, they become the propositions for the dataset. Possible TL specifications could be either ``person and car'' or ``person until car''. Applying this approach to autonomous driving, we similarly annotate the Waymo and NuScenes datasets, using detected ground truth object labels to create TL annotations. The total number of frames of TLV datasets is detailed in \cref{tab:dataset_stats}.

\begin{table}[ht]
  \centering
    \captionsetup{belowskip=-30pt}
    \begin{tabular}{lcccc}
    \toprule
    \textbf{Ground Truth TL Specifications} & \multicolumn{2}{c}{\textbf{Synthetic TLV Dataset}} & \multicolumn{2}{c}{\textbf{Real TLV Dataset}} \\
    \cmidrule(lr){2-3} \cmidrule(lr){4-5}
          & COCO  & ImageNet & Waymo & Nuscenes \\
    \midrule
    Eventually Event A & 15,750 & 15,750 & -     & - \\
    Always Event A & 15,750 & 15,750 & -     & - \\
    Event A And Event B & 31,500 & -     & -     & - \\
    Event A Until Event B & 15,750 & 15,750 & 8,736 & 19,808 \\
    (Event A And Event B) Until Event C & 31,500 & -     & 5,789 & 7,459 \\
    \bottomrule
    \end{tabular}%
    \caption{The number of frames in Synthetic TLV and Real TLV Datasets.}
  \label{tab:dataset_stats}%
\end{table}%

\subsection{Ground Truth Temporal Logic Specifications}
\label{subsec:tlv_spec}
\begin{enumerate}
    \item \textbf{Always/Eventually event A.} These temporal logic queries can be expressed as either $\Phi = \Box \hspace{1.0mm} p_k$ or $\diamondsuit \hspace{1.0mm} p_k$, identifying occurrences of a single event within a video. For instance, a natural language equivalent of this query is ``Identify all frames depicting a car accident'' or ``Locate all frames showcasing an animal's presence''. \label{prop:1}
    \item \textbf{Event A and B.} This scenario involves the concurrent occurrence of two distinct events streams, expressed as $\Phi = p_1 \wedge p_2$. Natural language specifications such as ``Locate all frames that have a pedestrian and a vehicle'' or ``Locate scenes that show a car accident and a pedestrian crossing'' can be represented by this TL query. \label{prop:2}
   \item\textbf{Event A until event B.} 
   It evaluates scenarios where a specific event $p_1$ is to be identified and monitored until another event $p_2$ occurs, expressed as $\Phi = p_1 \hspace{1.0mm} \mathsf{U} \hspace{1.0mm} p_2$. For instance, ``Identify all scenes where a pedestrian is walking until a vehicle passes by''. \label{prop:3}
    \item \textbf{Event A and B until C. }This specification encapsulates a video where two simultaneously occurring events are monitored up to the point a third event occurs; for example, ``Recognize scenes that show a child playing and dog barking until an adult appears''. It is expressed by $\Phi = (p_1 \wedge p_2) \hspace{1.0mm} \mathsf{U} \hspace{1.0mm} p_3$.  \label{prop:4}    
\end{enumerate}
The synthetic TLV dataset is constructed based on all the TL specifications mentioned above. In contrast, the AV datasets are more specifically focused on the TL specifications of ``Event A until B'' and ``Event A and B until C''. This is due to the fact that generating some TL specifications for real datasets would require altering the original video sequences, a scenario we aim to avoid.
\section{Experiment}
\label{sec:experiments}

In our experiments, we assess \coolname \ on diverse scene identification tasks using multiple datasets, aiming to answer key questions:

\begin{itemize}
    \item How significantly does the choice of a \textit{neural perception model} influence the performance of scene identification in the \coolname \ system?
    \item Given the optimal perception model, how does the efficacy of \coolname, which employs TL to reason over detected propositions per frame, compare to that of a Large Language Model (LLM) across various TL specifications?
    \item How does the duration of a video affect the performance of our method when the event is rare (in a long-duration video) or is temporally extended?
\end{itemize}

\subsection{Neural Perception Models} 
\label{sec:experiments:npm}
We utilize various neural perception models to assess the impact of different perception models on scene identification. We leverage SOTA computer vision models such as YOLO V8 \cite{YOLO}, Grounding Dino \cite{groundingdino}, Masked R-CNN \cite{mrcnn}, and Contrastive Language-Image Pre-training (CLIP) \cite{clip}. We use an image detection model instead of a video detection model for several reasons. First, we define a video as a sequence of frames, so detecting objects in each frame effectively means detecting objects throughout the entire video. Second, image detection can process per-frame input faster than video object detectors, making it essential for real-time video understanding, which is one of our key contributions.

\subsection{Benchmarks for Scene Identification}
Analysis of our method, benchmarked against video-language foundation models such as Video-Llama and ViCLIP, reveals their inherent limitations in handling long-term and complex temporal queries (see \cref{fig:fig2_nsvs_vs_benchmark_videomodels}). While these models perform well with shorter sequences, they lack robust temporal reasoning capabilities and struggle with extended video lengths. This limitation makes them less effective for long-term event detection tasks. Moreover, no current models exist that are capable of retrieving precise scenes or frames based on textual description or TL specification, despite their ability to understand the semantics of video content.
Hence, we design benchmarks for a balanced comparison that uses LLMs for reasoning over per-frame annotations instead of TL. Initially, neural perception models provide per-frame annotations to LLMs such as GPT. These LLMs are then prompted to identify scenes of interest according to specified propositions and TL specifications. We use these benchmarks to evaluate the impact of video length on scene identification performance. The prompts and methodologies applied in these comparisons are elaborated in the Appendix. 

\begin{table}[ht]
    \centering
    \captionsetup{belowskip=-30pt}
    \scriptsize 
    \begin{tabular}{@{}lccccccccc@{}}
        \toprule
        Model & \multicolumn{5}{c}{COCO-TLV + ImageNet-TLV} & \multicolumn{2}{c}{Waymo-TLV} & \multicolumn{2}{c}{NuScenes-TLV} \\
        \cmidrule(lr){2-6} \cmidrule(lr){7-8} \cmidrule(lr){9-10}
                       & \makecell{\(\diamondsuit\) A} & \makecell{\(\Box\) A} & \makecell{A \(\wedge\) B} & \makecell{A \(\mathsf{U}\) B} & \makecell{(A \(\wedge\) B) \(\mathsf{U}\) C} & \makecell{A \(\mathsf{U}\) B} & \makecell{(A \(\wedge\) B) \(\mathsf{U}\) C} & \makecell{A \(\mathsf{U}\) B} & \makecell{(A \(\wedge\) B) \(\mathsf{U}\) C} \\
        \midrule
        Grounding Dino & 0.30 & 0.50 & 0.28 & 0.30 & 0.33 & -    & -    & -    & -    \\
        CLIP-ViT-B-32 & 0.69 & 0.68 & 0.12 & 0.60 & 0.47 & -    & -    & -    & -    \\
        YOLOv8n       & 0.55 & 0.51 & 0.66 & 0.43 & 0.58 & 0.20 & 0.18 & 0.17 & 0.05 \\
        YOLOv8m       & 0.75 & 0.72 & 0.84 & 0.67 & 0.77 & 0.38 & 0.28 & 0.31 & 0.16 \\
        YOLOv8x       & 0.79 & 0.77 & \textbf{0.85} & 0.72 & \textbf{0.80} & 0.48 & 0.38 & 0.32 & 0.17 \\
        Masked R-CNN  & \textbf{0.81} & \textbf{0.84} & \textbf{0.85} & \textbf{0.75} & 0.74 & \textbf{0.83} & \textbf{0.80} & \textbf{0.64} & \textbf{0.52} \\
        \bottomrule
    \end{tabular}
    \caption{\textbf{Comprehensive Performance Comparison Across Models and TLV Datasets.} This illustrates the impact of different neural perception models across various datasets. The meaning of each symbol is described in  \cref{subsec:tlv_spec}.}
    \label{tab:performance_per_npm}
\end{table}

\begin{figure}[!t]
    \centering
    \subcaptionbox{\label{fig:fig5a_performance_different_nn}}
    {
      \includegraphics[width=0.48\textwidth]{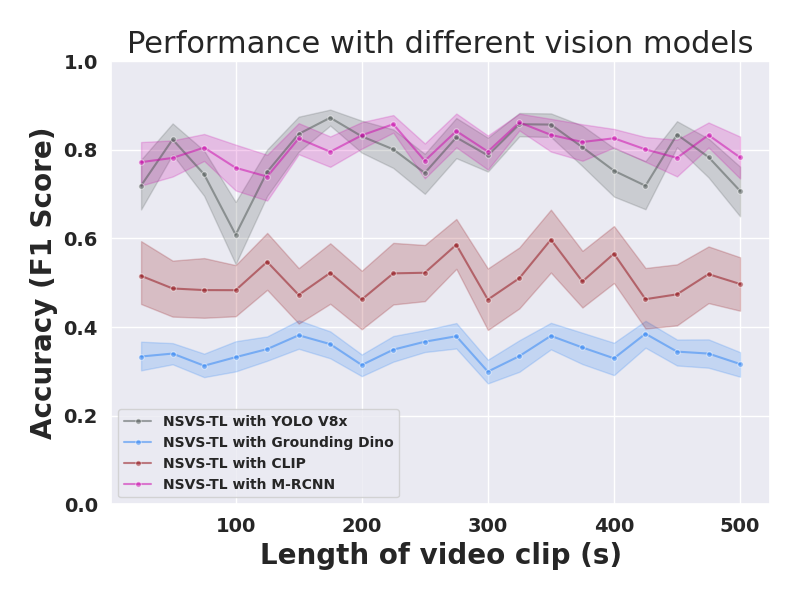}%
    } 
    \subcaptionbox{\label{fig:fig5b_performance_in_durations}}
    {
      \includegraphics[width=0.48\textwidth]{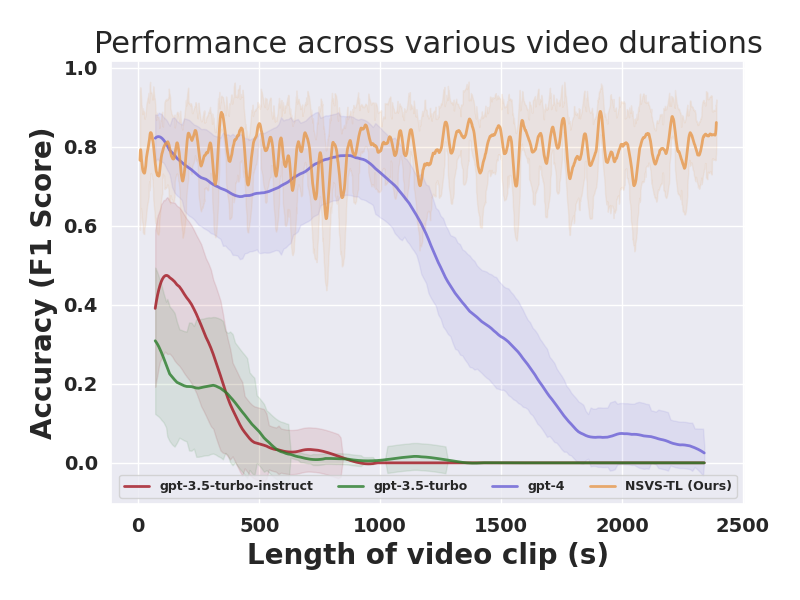}%
    }
    \caption{\textbf{Performance of \coolname \ Across Different Video Lengths.} \cref{fig:fig5a_performance_different_nn} demonstrates the impact of various neural perception models on scene identification performance. Additionally, \cref{fig:fig5b_performance_in_durations} illustrates the F1 scores for scene retrieval against the video length, fulfilling the ``A until B'' temporal specification.}
    \label{fig:fig5}
\end{figure}

\subsection{Evaluation Metrics}
We evaluate our method using multi-class classification metrics such as precision, recall, and F1 score for scene identification. The F1 score was used to evaluate our method with a balanced consideration of both precision (correctly identified frames / all frame identifications) and recall (correctly identified frames / correct ground truth frames), aiming to minimize false positives and false negatives. This measure is crucial for accurately detecting and retrieving relevant frames, indicating the effectiveness of our method in pinpointing specific scenes. The TLV Dataset, with its ground truth annotations, was used for this evaluation.

\section{Results}
\label{sec:results}

\begin{figure*}[t]
    \centering
    {
      \includegraphics[width=\textwidth]{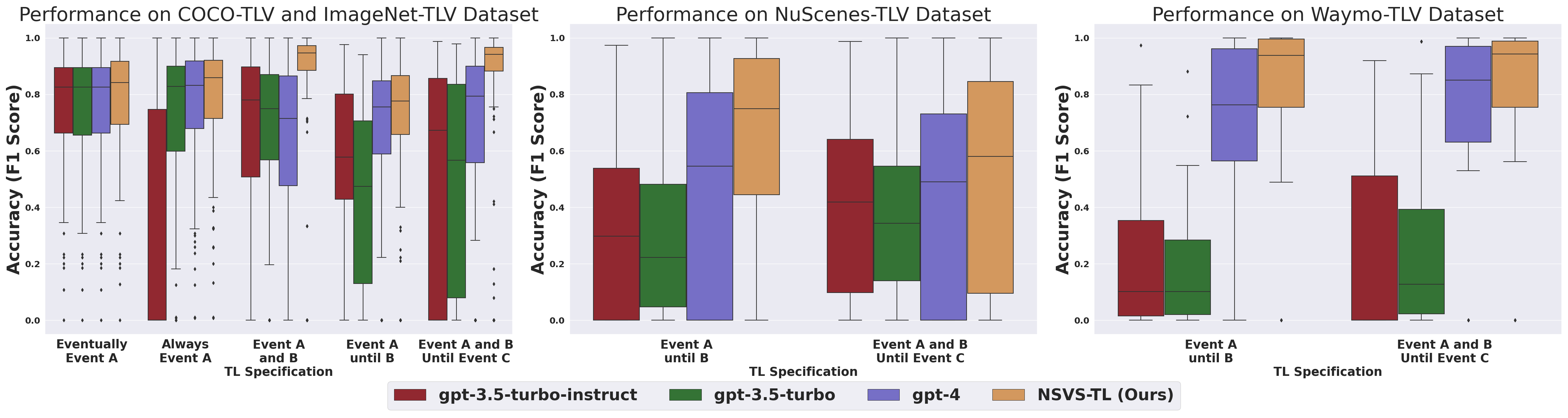}%
    } 
    {
      \includegraphics[width=\textwidth]{figures/fig7_result_with_graphic.pdf}%
    }
    \caption{\textbf{Comparative Performance of \coolname \ Across Complex Temporal Logic Specifications.} The upper box plots demonstrate \coolname's performance to benchmarks across different TL specifications. The lower section displays selected scenes from the NuScenes-TLV dataset, highlighting practical applications and results.}
    \label{fig:result_with_graphic}
\end{figure*}

\paragraph{Impact of different Neural Perception Models.}
Our work introduces a novel methodological approach that combines a neural perceptual model with symbolic methods. This integration provides a new perspective in understanding long-horizon videos. In fact, we propose a framework that allows for the integration of any neural perceptual models into our work, enhancing the capability to understand videos and enabling us to localize frames of interest with respect to queries. We show a comparison between different neural perceptual models in \cref{tab:performance_per_npm} and \cref{fig:fig5a_performance_different_nn}. We observe that \coolname \ with various neural perception models performs differently depending on the complexity of the TL specification and datasets.
The variation underscores the importance of these models' inherent visual detection capabilities for scene identification. 

\paragraph{Comparative Analysis of Temporal Logic and LLM-based Reasoning for Frame Retrieval.}
We evaluate TL-based reasoning against LLM-based reasoning over per-frame semantic annotations derived from neural perception models. The evaluation is conducted on four datasets: COCO-TLV, ImageNet-TLV, NuScenes-TLV, and Waymo-TLV. The results are collectively presented in \cref{fig:result_with_graphic}. For single event scenarios such as \textit{``Always/Eventually Event A Occurs''}, both our method and LLM-based reasoning perform reasonably well since these events do not require complex reasoning. In multi-event scenarios (\textit{``Event A and Event B'', ``Event A until Event B," and ``Event A and Event B until Event C"}), our TL-based reasoning outperforms all LLM-based baselines.

\paragraph{Impact of Video Length and Temporally Extended Scenarios on scene identification.}
We evaluate multi-event sequences with temporally extended gaps, such as those in the specification of events A until B. For these scenarios, we design a scene identification task that extends the video length: event A starts at the beginning and continues until event B occurs at the end. This tests the reasoning capacity of the LLM benchmarks as the temporal distance between events increases. The scene identification performance of GPT-3.5 and GPT-3.5 Turbo Instruct degrade within videos that are $500$ seconds long as illustrated in \cref{fig:fig5b_performance_in_durations}. GPT-4, on the other hand, suffers from a sharp decline in performance for videos whose lengths range from $1000$ seconds and more. In contrast, \(\coolname\) maintains consistent performance throughout. This consistency is observed even in videos spanning up to $40$ minutes or $2400$ seconds, indicating  \coolname's potential reliability in handling even longer videos.

\section{Conclusion}
\label{conclusion}
\coolname, a neuro-symbolic method, enhances video understanding and scene identification in contexts demanding extended temporal reasoning. \coolname, advances scene identification across various video durations by integrating semantic understanding with temporal reasoning. Unlike other methods focusing on short-duration video reasoning, we propose a novel approach for comprehending long-horizon, temporally extended videos. \\ \\
\textbf{Limitations and future work.}
\coolname's per-frame neural perception is limited in capturing multi-frame semantics, especially in scenarios that require understanding sequences, like ``man falling from a horse'', though it performs well with simpler scenarios such as ``man on a horse''. Future work will focus on interpreting multi-frame events, thereby extending its use in more complex video understanding tasks.
\section*{Acknowledgments}
\label{sec:acknowledgements}
This material is based upon work supported in part by the Office of Naval Research (ONR) under Grant No. N00014-22-1-2254 and N00014-24-1-2097. Additionally, this work was supported by the Defense Advanced Research Projects Agency (DARPA) contract FA8750-23-C-1018 and DARPA ANSR: RTX CW2231110. Approved for Public Release, Distribution Unlimited.
\clearpage
\section{Appendix}
\subsection{Examples of Temporal Logic}
\paragraph{Linear Temporal Logic (LTL).} 
In logic, linear temporal logic (LTL), also known as linear-time temporal logic, is one of the most commonly used temporal logic. LTL encodes formulae about the future of paths, e.g., a condition will eventually be true, a condition will be true until another fact becomes true, etc. LTL is composed of a set of atomic propositions (propositions that are either true or false and cannot be divided into multiple propositions), first-order propositional logic operators ($\wedge$ and, $\vee$ or, $\neg$ not, etc.), temporal operators ($\Box$ always, $\diamondsuit$ eventually, $\mathsf{X}$ next, $\mathsf{U}$ until, $\mathsf{R}$ release, etc.). The common temporal operators in LTL are
\begin{itemize}
    \item $\Box \phi$: $\phi$ is true on the entire subsequent path.
    \item $\diamondsuit \phi$: $\phi$ is true at least once on the subsequent path.
    \item $\mathsf{X} \phi$: $\phi$ is true at the next state.
    \item $\phi_1 \mathsf{U} \phi_2$: $\phi_1$ must be true until the first time $\phi_2$ is true, then $\phi_2$ must be true for all the future states.
    \item $\phi_1 \mathsf{R} \phi_2$: $\phi_1$ has to be true until and including the point where $\phi_2$ first becomes true.
\end{itemize}
The syntax of LTL is defined as
\begin{center}
    $
    \Phi ::= p \mid \neg \Phi \mid \Phi \vee \Phi \mid \mathsf{X}\Phi \mid \Phi\mathsf{U}\Phi
    $.
\end{center}

\paragraph{Computational Tree Logic (CTL).}
Computational tree logic (CTL) is a branching-time temporal logic. CTL's model of time is a tree structure in which the future is not determined. Many possible paths are branching out in the future, and one of the paths may turn into reality. CTL encodes formulae about all these branches of the future, i.e., something eventually will happen among all the possible futures, or something happens in at least one of the future branches. 

CTL shares the same set of first-order logic operators as LTL and includes all the temporal operators defined in the LTL section. In addition to the existing temporal operators, CTL includes two additional temporal quantifiers: All ($\mathsf{A}$) and Exists ($\mathsf{E}$). They are formally defined as:

\begin{itemize}
    \item $\mathsf{A} \phi$: $\phi$ is true on all paths starting from the current state.
    \item $\mathsf{E} \phi$: $\phi$ is true on at least one of the paths starting from the current state.
\end{itemize}

The syntax of CTL is formally defined as
\begin{center}
    $
    \Phi ::= p \mid \neg \Phi \mid \Phi \vee \Phi \mid \mathsf{E} \Box \Phi \mid \mathsf{EX}\Phi \mid \mathsf{E}[\Phi\mathsf{U}\Phi]
    $.
\end{center}

Note that compared to LTL, CTL can quantify multiple branches while LTL is capable of one path. However, LTL can encode formulas like "Eventually-Always something happens," while CTL is not capable of.

\paragraph{Probabilistic Computational Tree Logic (PCTL).}
Probabilistic CTL (PCTL) is an extension of CTL that allows for probabilistic quantification of described properties. The syntax and operations of PCTL are identical to CTL, except for the temporal quantifiers. Instead of the quantifiers (All and Exists) in CTL, PCTL uses probabilistic quantifiers $\mathsf{P}_{\sim \lambda}$. The syntax is formally defined as
\begin{center}
    $
    \Phi ::= p \mid \neg \Phi \mid \Phi \vee \Phi \mid \mathsf{P}_{\sim\lambda} [\Box \Phi] \mid \mathsf{P}_{\sim\lambda}[\Phi\mathsf{U}\Phi]
    $,
\end{center}
where $\sim \in \{<, \le, \ge, >\}$, and $\mathsf{P}_{\sim \lambda}$ indicates the required range of the probability of $\Phi$ being satisfied. For example, let $\sim = >$ and $\lambda = 0.6$, $\mathsf{P}_{> 0.6} \Phi$ is True if and only if $\Phi$ is satisfied over 60\% of the time, i.e., $\Phi$ has more than 60\% probability of being satisfied.

\paragraph{Linear Temporal Logic over Finite Traces.} 
LTL over infinite traces was introduced in the field of computer science as a specification language designed for concurrent programs \cite{Pnueli77LTL}. It facilitates the characterization of temporal reasoning within the domains of propositions and first-order logic. In our approach, frames of interest are considered as a finite set of frames extracted from a video. We can employ \textit{Linear temporal logic over finite traces} ($\text{LTL}_f$) to handle finite sequences encompassing finite-length trajectories, frames from video, system procedures. Given a set of atomic propositions $P$, the syntax of $\textit{LTL}_f$ formulas is defined as follows:

\begin{equation} 
    \label{eq:(3)}
    \phi ::= p \in P \mid \neg \phi \mid \phi_1 \wedge \phi_2 \mid \mathcal{X} \phi \mid \phi_1 \mathcal{U} \phi_2.
\end{equation} 

where $p \in P$ is an atomic proposition. It incorporates conditional operators such as AND ($\wedge$), OR ($\vee$), XOR($\oplus$), NOT($\neg$), IMPLY($\to$), along with temporal operators, ALWAYS($\Box$), EVENTUALLY($\diamondsuit$), NEXT($\mathcal{X}$), and UNTIL($\mathcal{U}$). 

\subsection{LLM Benchmark Prompts}
We design benchmarks for a balanced comparison that uses LLMs for reasoning over per-frame annotations instead of Temporal Logic (TL). Initially, neural perception models provide per-frame annotations to LLMs such as GPT. These LLMs are then prompted to identify scenes of interest according to specified propositions and TL specifications. We use these benchmarks to evaluate the impact of video length on scene identification performance. The prompts and methodologies applied in these comparisons are elaborated as follows:

\begin{tcolorbox}[
    title=Prompt for LLMs,
    colback=white,
    colframe=gray,
    colbacktitle=gray,
    sharp corners
]
    \ttfamily\scriptsize
    You are a Language Assistant to enable searching a video frame for objects satisfying a condition.
    
    TASK DESCRIPTION:
    I will be giving you the list of objects detected in the video and a condition from the search. The list of objects will be specified in this format and include those frames where an object is detected:
    
     - Frame 1: [object1, object2, object3]

     - Frame 2: [object1, object2]
               
    A condition is of the form: 
    
    - object1 Until object2,
    
    - object1 Until (object2 and object3),
    
    - Eventually object 1,
    
    - Always object 2, etc.

    You can use the list of objects detected in the video to help you. You don't have access to the video. Your job is to specify the frame numbers where the rule is satisfied. There can be many such sequences. If you cannot find any sequence, you must return output [None]. To clarify, you will be looking for frames where the rule is satisfied. The output is a list of frame numbers, such as [1,20, 25, 50], specifying the frames where the rule holds. An example of the list of detections per frame is as follows:

    - 01: [person, car]
    
    - 02: [person, car]
    
    - 10: [person, car]
    
    - 30: [truck]
    
    - 40: [car]
    
    - 44: [person]
    
    - 50: [car, truck]
    
    Case 1: `person' until `truck' gives all frames of `person' and `truck' if `truck' shows up. \\ 
    
    Case 2: Eventually `person' gives all frames of `person' eventually showing up. \\

    Case 3: `car' and `truck' should give all frames of `car', and `truck' showing up in the same frame. \\
    
    Case 4: Always `car' should give all frames of `car' always showing up. \\

    Case 5: (`car' and `truck') until `person' gives all frames of `car' and `truck' showing up in the same frame if 'person' shows up; all frames of 'person' included.
    
\end{tcolorbox}

\bibliographystyle{splncs04}
\bibliography{ref}

\begin{thebibliography}{10}
\providecommand{\url}[1]{\texttt{#1}}
\providecommand{\urlprefix}{URL }
\providecommand{\doi}[1]{https://doi.org/#1}

\bibitem{BertasiusWT21}
Bertasius, G., Wang, H., Torresani, L.: Is space-time attention all you need for video understanding? In: Meila, M., Zhang, T. (eds.) International Conference on Machine Learning. Proceedings of Machine Learning Research, vol.~139, pp. 813--824. {PMLR} (2021)

\bibitem{temporal-event-detect}
Cheng, Y., Fan, Q., Pankanti, S., Choudhary, A.N.: Temporal sequence modeling for video event detection. In: {IEEE} Conference on Computer Vision and Pattern Recognition. pp. 2235--2242. {IEEE} Computer Society, Columbus, OH, USA (2014)

\bibitem{imagenet}
Deng, J., Dong, W., Socher, R., Li, L.J., Li, K., Fei-Fei, L.: Imagenet: A large-scale hierarchical image database. pp. 248--255 (2009). \doi{10.1109/CVPR.2009.5206848}

\bibitem{interpretable-anomaly-detect}
Doshi, K., Yilmaz, Y.: Towards interpretable video anomaly detection. In: {IEEE/CVF} Winter Conference on Applications of Computer Vision. pp. 2654--2663. {IEEE}, Waikoloa, HI, USA (2023)

\bibitem{Temporal-and-Modal-Logic}
Emerson, E.A.: Temporal and modal logic. In: Handbook of Theoretical Computer Science, Volume B: Formal Models and Sematics (1991), \url{https://api.semanticscholar.org/CorpusID:6062082}

\bibitem{short1}
Feichtenhofer, C., Fan, H., Malik, J., He, K.: Slowfast networks for video recognition. In: {IEEE/CVF} International Conference on Computer Vision. pp. 6201--6210. {IEEE} (2019)

\bibitem{action-detect-2}
Feichtenhofer, C., Pinz, A., Wildes, R.P.: Spatiotemporal multiplier networks for video action recognition. In: {IEEE} Conference on Computer Vision and Pattern Recognition. pp. 7445--7454. {IEEE} Computer Society, Honolulu, HI, USA (2017)

\bibitem{action-detect}
Feichtenhofer, C., Pinz, A., Zisserman, A.: Convolutional two-stream network fusion for video action recognition. In: {IEEE} Conference on Computer Vision and Pattern Recognition. pp. 1933--1941. {IEEE} Computer Society, Las Vegas, NV, USA (2016)

\bibitem{nuscenes}
Guo, Y., Caesar, H., Oscar, B., Philion, J., Fidler, S.: The efficacy of neural planning metrics: A meta-analysis of pkl on nuscenes. In: IROS 2020 Workshop on Benchmarking Progress in Autonomous Driving (2020)

\bibitem{mrcnn}
He, K., Gkioxari, G., Dollár, P., Girshick, R.: Mask r-cnn (2017)

\bibitem{long2}
Huang, Q., Xiong, Y., Rao, A., Wang, J., Lin, D.: Movienet: {A} holistic dataset for movie understanding. In: Vedaldi, A., Bischof, H., Brox, T., Frahm, J. (eds.) European Conference on Computer Vision. Lecture Notes in Computer Science, vol. 12349, pp. 709--727. Springer (2020)

\bibitem{event-detect-spatiotemporal}
Jiang, F., Yuan, J., Tsaftaris, S.A., Katsaggelos, A.K.: Anomalous video event detection using spatiotemporal context. Comput. Vis. Image Underst.  \textbf{115}(3),  323--333 (2011). \doi{10.1016/J.CVIU.2010.10.008}, \url{https://doi.org/10.1016/j.cviu.2010.10.008}

\bibitem{junges2021stormpy}
Junges, S., Volk, M.: Stormpy - python bindings for storm (2021)

\bibitem{stormpy}
Junges, S., Volk, M.: Stormpy - python bindings for storm (2021), \url{github.com/moves-rwth/stormpy}

\bibitem{neural-symbolic-cv}
Kroshchanka, A., Golovko, V., Mikhno, E., Kovalev, M., Zahariev, V., Zagorskij, A.: A neural-symbolic approach to computer vision. In: International Conference on Open Semantic Technologies for Intelligent Systems. pp. 282--309. Springer (2021)

\bibitem{lstm-event-detect}
Li, N., Chang, F., Liu, C.: Human-related anomalous event detection via spatial-temporal graph convolutional autoencoder with embedded long short-term memory network. Neurocomputing  \textbf{490},  482--494 (2022)

\bibitem{coco}
Lin, T.Y., Maire, M., Belongie, S., Bourdev, L., Girshick, R., Hays, J., Perona, P., Ramanan, D., Zitnick, C.L., Dollár, P.: Microsoft coco: Common objects in context (2014)

\bibitem{maaz2023video}
Maaz, M., Rasheed, H., Khan, S., Khan, F.S.: Video-chatgpt: Towards detailed video understanding via large vision and language models. arXiv preprint arXiv:2306.05424  (2023)

\bibitem{Manna}
Manna, Z., Pnueli, A.: The Temporal Logic of Reactive and Concurrent Systems: Specification. Springer-Verlag (1992)

\bibitem{visual-symbolic}
Mavroudi, E., Haro, B.B., Vidal, R.: Representation learning on visual-symbolic graphs for video understanding. In: European Conference on Computer Vision. Lecture Notes in Computer Science, vol. 12374, pp. 71--90. Springer (2020)

\bibitem{event-detection-video-stream}
Medioni, G.G., Cohen, I., Br{\'{e}}mond, F., Hongeng, S., Nevatia, R.: Event detection and analysis from video streams. {IEEE} Trans. Pattern Anal. Mach. Intell.  \textbf{23}(8),  873--889 (2001)

\bibitem{openai2023gpt4}
OpenAI: Gpt-4 technical report. arXiv preprint arXiv:2303.08774  (2023)

\bibitem{Pnueli77LTL}
Pnueli, A.: The temporal logic of programs. In: Symposium on Foundations of Computer Science. pp. 46--57 (1977). \doi{10.1109/SFCS.1977.32}, \url{https://doi.org/10.1109/SFCS.1977.32}

\bibitem{clip}
Radford, A., Kim, J.W., Hallacy, C., Ramesh, A., Goh, G., Agarwal, S., Sastry, G., Askell, A., Mishkin, P., Clark, J., Krueger, G., Sutskever, I.: Learning transferable visual models from natural language supervision (2021)

\bibitem{YOLO}
Redmon, J., Divvala, S., Girshick, R., Farhadi, A.: You only look once: Unified, real-time object detection (2016)

\bibitem{symbolic-high-speed-video}
Sarkar, S., Lore, K.G., Sarkar, S.: Early detection of combustion instability by neural-symbolic analysis on hi-speed video. In: Proceedings of the {NIPS} Workshop on Cognitive Computation: Integrating Neural and Symbolic Approaches co-located with the 29th Annual Conference on Neural Information Processing Systems. {CEUR} Workshop Proceedings, vol.~1583. CEUR-WS.org, Montreal, Canada (2015)

\bibitem{securitycams}
SecurityInfoWatch: Data generated by new surveillance cameras to increase exponentially in the coming years (2016), \url{https://www.securityinfowatch.com/video-surveillance/news/12160483/data-generated-by-new-surveillance-cameras-to-increase-exponentially-in-the-coming-years}, accessed: 2023-09-30

\bibitem{waymo}
Sun, P., Kretzschmar, H., Dotiwalla, X., Chouard, A., Patnaik, V., Tsui, P., Guo, J., Zhou, Y., Chai, Y., Caine, B., Vasudevan, V., Han, W., Ngiam, J., Zhao, H., Timofeev, A., Ettinger, S., Krivokon, M., Gao, A., Joshi, A., Zhang, Y., Shlens, J., Chen, Z., Anguelov, D.: Scalability in perception for autonomous driving: Waymo open dataset (June 2020)

\bibitem{long1}
Tapaswi, M., Zhu, Y., Stiefelhagen, R., Torralba, A., Urtasun, R., Fidler, S.: Movieqa: Understanding stories in movies through question-answering. In: {IEEE} Conference on Computer Vision and Pattern Recognition. pp. 4631--4640. {IEEE} Computer Society (2016)

\bibitem{touvron2023llama}
Touvron, H., Martin, L., Stone, K., Albert, P., Almahairi, A., Babaei, Y., Bashlykov, N., Batra, S., Bhargava, P., Bhosale, S., et~al.: Llama 2: Open foundation and fine-tuned chat models. arXiv preprint arXiv:2307.09288  (2023)

\bibitem{short2}
Tran, D., Wang, H., Feiszli, M., Torresani, L.: Video classification with channel-separated convolutional networks. In: {IEEE/CVF} International Conference on Computer Vision. pp. 5551--5560. {IEEE} (2019)

\bibitem{long-form-video}
Wu, C., Kr{\"{a}}henb{\"{u}}hl, P.: Towards long-form video understanding. In: {IEEE} Conference on Computer Vision and Pattern Recognition. pp. 1884--1894. Computer Vision Foundation / {IEEE} (2021)

\bibitem{long3}
Xiong, Y., Huang, Q., Guo, L., Zhou, H., Zhou, B., Lin, D.: A graph-based framework to bridge movies and synopses. In: {IEEE/CVF} International Conference on Computer Vision. pp. 4591--4600. {IEEE} (2019)

\bibitem{cnn-event-detect}
Xu, Z., Yang, Y., Hauptmann, A.G.: A discriminative {CNN} video representation for event detection. In: {IEEE} Conference on Computer Vision and Pattern Recognition. pp. 1798--1807. {IEEE} Computer Society, Boston, MA, USA (2015)

\bibitem{event-extract}
Yang, G., Li, M., Zhang, J., Lin, X., Ji, H., Chang, S.: Video event extraction via tracking visual states of arguments. In: Williams, B., Chen, Y., Neville, J. (eds.) {AAAI} Conference on Artificial Intelligence. pp. 3136--3144. {AAAI} Press, Washington, DC, USA (2023)

\bibitem{yang2023specification}
Yang, Y., Gaglione, J.R., Chinchali, S., Topcu, U.: Specification-driven video search via foundation models and formal verification. arXiv preprint arXiv:2309.10171  (2023)

\bibitem{neural-symbolic}
Yi, K., Wu, J., Gan, C., Torralba, A., Kohli, P., Tenenbaum, J.: Neural-symbolic {VQA:} disentangling reasoning from vision and language understanding. In: Bengio, S., Wallach, H.M., Larochelle, H., Grauman, K., Cesa{-}Bianchi, N., Garnett, R. (eds.) Advances in Neural Information Processing Systems. pp. 1039--1050 (2018)

\bibitem{graphical-model-relationship-detection}
Yu, D., Yang, B., Wei, Q., Li, A., Pan, S.: A probabilistic graphical model based on neural-symbolic reasoning for visual relationship detection. In: {IEEE/CVF} Conference on Computer Vision and Pattern Recognition. pp. 10599--10608. {IEEE}, New Orleans, LA, USA (2022)

\bibitem{groundingdino}
and Zeng, Z., Ren, T., Li, F., Zhang, H., Yang, J., Li, C., Yang, J., Su, H., Zhu, J., et~al.: Grounding dino: Marrying dino with grounded pre-training for open-set object detection. arXiv preprint arXiv:2303.05499  (2023)

\bibitem{damonlpsg2023videollama}
Zhang, H., Li, X., Bing, L.: Video-llama: An instruction-tuned audio-visual language model for video understanding. arXiv preprint arXiv:2306.02858  (2023), \url{https://arxiv.org/abs/2306.02858}

\bibitem{zheng2022abnormal}
Zheng, X., Zhang, Y., Zheng, Y., Luo, F., Lu, X.: Abnormal event detection by a weakly supervised temporal attention network. CAAI Transactions on Intelligence Technology  \textbf{7}(3),  419--431 (2022)

\end{thebibliography}
\end{document}